\documentclass[runningheads]{llncs}
\usepackage{graphicx}

\usepackage{tikz}
\usepackage{hyperref}
\usepackage{comment}
\usepackage{amsmath,amssymb} 
\usepackage{color}
\usepackage{wrapfig}
\usepackage[accsupp]{axessibility}  

\usepackage{booktabs}
\usepackage{multicol}
\usepackage{bbm}
\usepackage{multirow}
\usepackage{colortbl}
\usepackage{bbm}
\usepackage{amssymb}
\usepackage{bbold}
\usepackage{bm}
\usepackage{color}
\definecolor{mygray}{gray}{.9}
\definecolor{mycyan}{cmyk}{.3,0,0,0}
\definecolor{light-gray}{gray}{0.5}
\definecolor{c0}{RGB}{146,208,80}
\definecolor{c1}{RGB}{0,176,80}
\definecolor{c2}{RGB}{102,153,0} 
\definecolor{c3}{RGB}{0,0,255}
\definecolor{citationcolor}{RGB}{0,113,188}
\definecolor{ACM-purple}{RGB}{121,29,125}

\begin{document}
\pagestyle{headings}
\mainmatter
\def\ECCVSubNumber{59}  

\title{Contrastive Deep Supervision} 


\titlerunning{Contrastive Deep Supervision}
%
\author{Linfeng Zhang$^1$, Xin Chen$^2$, Junbo Zhang$^1$, Runpei Dong$^3$, Kaisheng Ma$^1$\thanks{Corresponding author}}
\institute{Tsinghua University$^1$, Intel Corporation$^2$, Xi'an Jiaotong University$^3$ \\ \texttt{zhang-lf19@mails.tsinghua.edu.cn}}

\authorrunning{L. Zhang et al.}
%
\maketitle

\begin{abstract}
  The success of deep learning is usually accompanied by the growth in neural network depth.
  However, the traditional training method only
  supervises the neural network at its last layer and propagates the supervision layer-by-layer, which leads to hardship in optimizing the intermediate layers.
  Recently, deep supervision has been proposed to add auxiliary classifiers to the intermediate layers of deep neural networks. By optimizing these auxiliary classifiers with the supervised task loss, the supervision can be applied to the shallow layers directly. However, deep supervision conflicts with the well-known observation that the shallow layers learn low-level features instead of task-biased high-level semantic features. To address this issue, this paper proposes a novel training framework named Contrastive Deep Supervision, which supervises the intermediate layers with augmentation-based contrastive learning. 
  Experimental results on nine popular datasets with eleven models demonstrate its effects on general image classification, fine-grained image classification and object detection in supervised learning, semi-supervised learning and knowledge distillation. Codes have been released in \href{https://github.com/ArchipLab-LinfengZhang/contrastive-deep-supervision}{\textcolor{red}{Github}}.
\end{abstract}

\section{Introduction}
\begin{figure}[t!]
    \begin{center}
        \includegraphics[width=11.0cm]{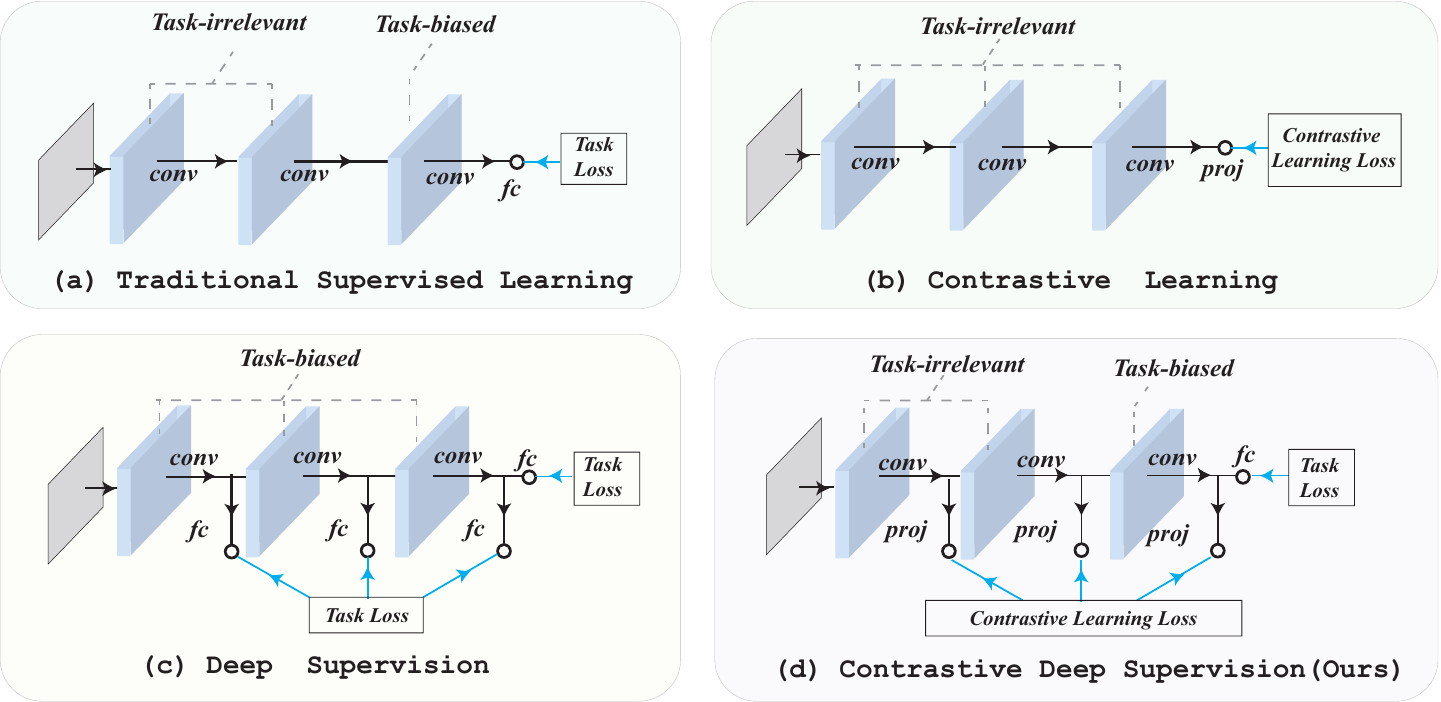}
    \end{center}
    \caption{\label{fig1} The overview of the four methods. ``$\rightarrow$'' and
    {``\textcolor[RGB]{31,163,221}{$\rightarrow$}'' } 
    indicate the path of forward computation and gradients backward computation. ``\texttt{proj}'' and ``\texttt{fc}'' indicate the projection heads and the fully connected classifiers, respectively. The gray dash line indicates whether the feature is task-irrelevant or task-biased. (a) Traditional supervised learning only applies supervision to the last layer and propagates it to the previous layers, leading to gradient vanishing. (c) Deep supervision trains both the last layer and the intermediate layers directly, which addresses gradient vanishing but makes all the layers be biased to the task. (d) Our method introduces contrastive learning to supervise the intermediate layer and thus avoid these problems. }
  \end{figure}

Along with the growth in large-scale datasets and computation resources, deep neural networks have become the most dominant models for various tasks~\cite{bert,fasterrcnn}. However, the increasing depth of neural networks also introduces challenges in their training process.
Traditional supervised training method only applies the supervision to the last layer and then propagates the error from the last layer to the shallow layers (Figure~\ref{fig1}(a)), which leads to hardship in optimizing the intermediate layers such as gradient vanishing~\cite{droplayer}. 

Recently, deep supervision (\emph{a.k.a. deeply-supervised net}) has been proposed to address this issue by optimizing the intermediate layers directly~\cite{deeplysupervisednet}. As shown in Figure~\ref{fig1}(b), deep supervision adds several auxiliary classifiers to the intermediate layers in different depths. During the training phase, these 
classifiers are optimized with the original final classifier together by the same training loss (\emph{e.g.} cross entropy for classification tasks). Both experimental and theoretical analyses have demonstrated its effectiveness in facilitating model convergence~\cite{wang2015training}.

However, success comes with remaining obstacles.  
In general, different layers in convolutional neural networks tend to learn features at different levels. Usually, the shallow layers learn low-level features such as colors and edges, while the last several layers learn more high-level task-related semantic features such as categorical knowledge for classification tasks~\cite{zhou2014object}.
However, deep supervision forces the shallow layers to learn the task-related knowledge, which disobeys the original feature extraction process in neural networks. As pointed out in MSDNet~\cite{multiscaledensenet}, this conflict sometimes leads to accuracy degradation in the final classifier. This observation indicates that the supervised task loss is probably not the best supervision for optimizing the intermediate layers.


In this paper, we argue that \emph{contrastive learning can provide better supervision for intermediate layers than the supervised task loss.}
Contrastive learning is one of the most popular and effective techniques in representation learning~\cite{simclr,simclr2,sup}.
Usually, it regards two augmentations from the same image as a positive pair and different images as negative pairs. In the training period, the neural network is trained to minimize the distance of a positive pair while maximizing the distance of a negative pair. As a result, the network can learn the invariance to various data augmentation, such as { \texttt{Color Jitter}} and { \texttt{Random Gray Scale}}. Considering that these data augmentation invariances are usually low-level, task-irrelevant and transferable to various vision tasks~\cite{contrast_seg,xie2021detco}, we argue that they are more beneficial knowledge to be learned by intermediate layers.

Motivated by these observations, we propose a novel training framework named \emph{Contrastive Deep Supervision}. It optimizes the intermediate layers with contrastive learning instead of traditional supervised learning. As shown in Figure~\ref{fig1}(d), several projection heads are attached in the intermediate layers of the neural networks and trained to perform contrastive learning. These projection heads can be discarded in the inference period to avoid additional computation and storage. 
Different from deep supervision which  trains the intermediate layers to learn the knowledge for a specific task, the intermediate layers in our method are trained to learn the invariance to data augmentation, which makes the neural network generalize better. Besides, since contrastive learning can be performed on unlabeled data, the proposed contrastive deep supervision can also be easily extended in the semi-supervised learning paradigm.

Moreover, contrastive deep supervision can be further utilized to boost the performance of another deep learning technique -- knowledge distillation. Knowledge distillation (KD) is a popular model compression approach which aims to transfer the knowledge from a cumbersome teacher model to a lightweight student model~\cite{model_compression,distill_hinton,bornagain}. Recently, abundant research finds that distilling the ``crucial knowledge'' inside the backbone features such as attention and relation~\cite{attentiondistillation,relational_kd,relational_kd2} leads to better performance than directly distilling all the backbone features. 
In this paper, we show that the data augmentation invariances learned by the intermediate layers in contrastive deep supervision are more beneficial knowledge to be distilled. By combining contrastive deep supervision with the na\"ive feature distillation, the distilled ResNet18 achieves 73.23\% accuracy on ImageNet,  which outperforms the baseline and the second-best KD method by 4.02\% and 2.16\%, respectively.

Extensive experiments on nine datasets with eleven neural networks methods have been conducted to evaluate its effectiveness on general image classification, fine-grained image classification, object detection in supervised learning, semi-supervised learning and knowledge distillation, which demonstrates that contrastive deep supervision enables neural networks to learn better visual representation.
In the discussion section, we further explain the effectiveness of our method from the perspective of regularization methods, which prevents models from overfitting and leads to better uncertainty estimation. 
To sum up, the main contributions of our paper can be summarized as follows.

\begin{itemize}
    \item We propose \textit{contrastive deep supervision}, a neural network training method in which the intermediate layers are directly optimized with contrastive learning. It enables neural networks to learn better visual representation at no expense of additional parameters and computation during inference.
    
    \item From the perspective of deep supervision, this paper firstly shows that the intermediate layers can be trained with supervision besides the task loss. 
    \item From the perspective of representation learning, we firstly show that contrastive learning and supervised learning can be combined in a one-stage deep-supervision manner instead of the two-stage ``pretrain-finetune'' scheme.
    \item Extensive experiments on nine datasets, eleven neural networks with eleven comparison methods demonstrate the effectiveness of our method on general classification, fine-grained classification and object detection in supervised learning, semi-supervised learning and knowledge distillation.
    
    

    
\end{itemize}

\section{Related Work}

\subsection{Deep Supervision}
 
Deep neural networks usually contain a large number of layers, which increases the difficulty of optimization.
To address this issue, deeply supervised net (\emph{a.k.a.} deep supervision) is proposed to directly supervise the intermediate layers of deep neural networks~\cite{deeplysupervisednet}. Wang~\emph{et~al.} show that deep supervision can alleviate the vanishing gradient problem and thus leads to significant performance improvements~\cite{wang2015training}. Usually, deep supervision attaches several auxiliary classifiers at the intermediate layers and supervises these auxiliary classifiers with the task loss (\emph{e.g.} cross-entropy loss in classification). 
Recently, several methods have been proposed to improve deep supervision with knowledge distillation, which aims to minimize the difference between the prediction of the deepest classifier and the auxiliary classifiers in the intermediate layers~\cite{dsk,li2020dynamic}. 
Besides classification, abundant research has also demonstrated the effectiveness of deep supervision methods in dynamic neural networks~\cite{Zhang2019SCANAS}, semantic segmentation~\cite{semantic_dsn,semantic_dsn2,DBLP:conf/cvpr/ReissSFRS21}, object detection~\cite{detect_dsn}, knowledge distillation~\cite{tofd} and so on.

\subsection{Contrastive Learning}
In the last several years, contrastive learning has become the most popular method in representation learning~\cite{DBLP:conf/cvpr/0010KBLY21,DBLP:conf/cvpr/00230W0W21,DBLP:conf/cvpr/JeonMKS21,DBLP:conf/cvpr/Hu00Q21,DBLP:conf/cvpr/YangL0L0021,DBLP:conf/cvpr/HanF0021,DBLP:conf/cvpr/WangL21a,DBLP:conf/cvpr/00010G0HC21,DBLP:conf/cvpr/HouGNX21,DBLP:conf/cvpr/WangHLXY021}.
Oord~\emph{et~al.} propose the contrastive predictive coding, which aims to predict the low dimension embedding of future signals with an auto-regressive model~\cite{cpc}.
He~\emph{et al.} propose MoCo, which introduces a dynamic memory bank to record the embeddings of negative samples~\cite{moco,mocov2,chen2021empirical}.
Then, SimCLR is proposed to show the importance of large batch size and long training time in contrastive learning~\cite{simclr,simclr2}.
Recently, abundant research has been proposed to study the influence of negative samples further.
BYOL is introduced to demonstrate that contrastive learning is effective even without negative samples~\cite{byol}. SimSiam gives a detailed study on the importance of batch normalization, negative samples, memory bank, and the stop-gradient operation~\cite{chen2020exploring}.
Besides self-supervised learning, contrastive learning has also shown its power in the traditional supervised learning paradigm. 
Khosla ~\emph{et al.} show that state-of-the-art performance can be achieved on ImageNet with the basic contrastive learning in SimCLR by building the positive pairs with label supervision~\cite{sup,chen2021wasserstein}. 
Park~\emph{et al.} apply contrastive learning to unpaired image-to-image translation, which breaks the limitation of cycle reconstruction~\cite{park2020contrastive}.

\subsection{Knowledge Distillation}
Knowledge distillation, which aims to facilitate the training of a lightweight student model under the supervision of an over-parameterized teacher model, 
has become one of the most popular methods in model compression.
Knowledge distillation is first proposed by Bucilua~\emph{et al.}~\cite{model_compression} and then expanded by Hinton~\emph{et al.}~\cite{distill_hinton},
who introduces a temperature-characterized softmax to soften the distribution of teacher logits.
Instead of distilling the knowledge of the logits, more and more techniques are proposed to distill the information in teacher features or its variants, such as
attention maps~\cite{attentiondistillation,liu2019paying}, negative values~\cite{kd_comprehensive}, task-oriented information~\cite{tofd}, relational information~\cite{relational_kd,relational_kd2,structuredkd}, Gram matrix~\cite{fsp_kd}, mutual information~\cite{kd_variational}, context information~\cite{detectiondistillation} and so on.
Besides model compression, knowledge distillation has also achieved significant success in self-supervised learning~\cite{kd_self1,kd_self2}, semi-supervised learning~\cite{kd_semi,kd_semi2}, multi-exit neural network~\cite{Zhang2019SCANAS,selfdistillation,slimmablev2}, incremental learning~\cite{kd_incremental} and
model robustness~\cite{noisy_students,auxiliarytraining}

\section{Methodology}
\subsection{Deep Supervision}
In this subsection, we revisit the formulation of deep supervision methods.
Let $c$ be a given backbone classifier, deep supervision introduces several shallow classifiers by using the intermediate features in $c$. More specifically, assume $c= g \circ f$ where $g$ is the final classifier, $f$ is the feature extractor operator and $f=f_K \circ f_{K-1}\circ \cdot \cdot \cdot f_1$. $K$ denotes the number of convolutional stages in $f$. 
At each feature extraction stage $i$, deep supervision attaches an auxiliary classifier $g_i$ for providing intermediate supervision. Thus, there are $K$ classifiers in total which have the following form:
\begin{equation}
    \begin{aligned}
    c_1(x)&=g_1 \circ f_1(x) \\
    c_2(x)&=g_2 \circ f_2 \circ f_1(x)\\
    \cdot \cdot \cdot\\
    c_K(x)&=g_K \circ f_K \circ f_{K-1} \circ \cdot \cdot \cdot \circ f_1(x).
    \end{aligned}
\end{equation}
Given a set of training samples $\mathcal{X}=\{x_i\}^n_{i=1}$ and its corresponding labels $\mathcal{Y}=\{y_i\}^n_{i=1}$, the training loss of deep supervision $\mathcal{L}_{\text{DS}}$ can be formulated as 
\begin{equation}
\label{loss:ds}
    \mathcal{L}_{\text{DS}}= \underbrace{\mathcal{L}_{\text{CE}}(c_K(\mathcal{X}), \mathcal{Y})}_{\text{from standard training}} + \alpha \cdot \mathop{\sum}_{i=1}^{K-1} \underbrace{\mathcal{L}_{\text{CE}}(c_i(\mathcal{X}), \mathcal{Y})}_{\text{from deep supervision}},
\end{equation}
where $\mathcal{L}_{\text{CE}}$ indicates the cross entropy loss. The first and the second item in the loss function indicate the standard training loss and the additional loss from deep supervision for the intermediate layers, respectively. $\alpha$ is a hyper-parameter to balance the two loss items. 
Recently, some research has been proposed to apply layer-wise consistency on deep supervision, which additionally minimizes the KL divergence between the prediction of auxiliary classifiers and the final classifier~\cite{dsk,li2020dynamic}. These methods can also be considered as the knowledge distillation which regards the final classifier as the teacher and the auxiliary classifiers as the students.
Their training loss can be formulated as 
\begin{equation}
\label{loss:dis}
 \mathcal{L}_{\text{DS}} + \beta \cdot \mathop{\sum}_{i=1}^{K-1} \mathcal{L}_{\text{KL}}(c_i(\mathcal{X}), c_K(\mathcal{X})),
\end{equation}
where $\beta$ is a hyper-parameter to balance the two loss functions.

\subsection{Contrastive Deep Supervision}
In this subsection, we first introduce the formulation of contrastive learning.
For a minibatch of $N$ images $\{x_1, x_2, ..., x_N\}$, we apply stochastic data augmentation to each image twice, resulting in a batch of $2N$ images. For convenience, we denote $x_i$  and $x_{N+i}$ images as the two augmentations from the same image, which is regarded as a positive pair. 
Denote $z= c(x)$ as the normalized projection head outputs, contrastive learning loss (\emph{a.k.a.} NT-Xtent~\cite{simclr}) can be formulated as  
\begin{equation}
\label{loss:contra}
    \mathcal{L}_{\text{Contra}}= - \sum_{i=1}^{N}\log  \frac{\exp(z_i\cdot z_{i+N})/\mathcal{\tau}}{\sum_{k=1}^{2N}\mathbb{1}_{[k\neq i]}\exp(z_i\cdot z_k)/\mathcal{\tau}},
\end{equation}
where $\mathbb{1}\in \{0, 1\}$ is an indicator function evaluating to 1 if $k\neq i$ and $\tau$ is a temperature hyper-parameter. Intuitively, $\mathcal{L}_{\text{Contra}}$ encourages the encoder network to learn similar representation for different augmentations from the same image while increasing the difference between representations of the augmentations from different images.

The main difference between deep supervision and our method is that deep supervision trains the auxiliary classifiers by the cross entropy loss while our method trains them with the contrastive loss $\mathcal{L}_{\text{Contra}}$. By denoting the contrastive loss at $c_i$ as $\mathcal{L}_{\text{Contra}}(\mathcal{X};c_i)$, then the training loss of our contrastive deep supervision $\mathcal{L}_{\text{CDS}}$ can be formulated as 
\begin{equation}
\label{loss:ours}
    \mathcal{L}_{\text{CDS}}= \underbrace{\mathcal{L}_{\text{CE}}(c_K(\mathcal{X}), \mathcal{Y})}_{\text{from standard training}} + \lambda_1 \mathop{\sum}_{i=1}^{K-1} \underbrace{\mathcal{L}_{\text{Contra}}(\mathcal{X};c_i)}_{\text{from our method}} ,
\end{equation}
where the first and the second item indicate the standard training loss and the additional loss in our method for the intermediate layers, respectively. $\lambda_1$ is a hyper-parameter to balance the two loss items. 

Based on the above formulation on supervised learning, we can extend contrastive deep supervision in semi-supervised learning and knowledge distillation.

\subsubsection{Semi-supervised Learning} 
In semi-supervised learning, we assume that there is a labeled dataset $\mathcal{X}_1$ with its labels $\mathcal{Y}_1$ and an unlabeled dataset $\mathcal{X}_2$. On the labeled data, contrative deep supervision can be appled directly with $\mathcal{L}_{\text{CDS}}$. On the unlabeled data, due to the lack of labels,  contrastive deep supervision only optimize the contrastive learning loss $\mathcal{L_{\text{Contra}}}$, which can be formulated as \begin{equation}
    \mathcal{L}_{\text{CDS}}(\mathcal{X}_1,\mathcal{Y}_1) + \mathcal{L}_{\text{Contra}}(\mathcal{X}_2)
\end{equation} 

\subsubsection{Knowledge Distillation}



The intermediate layers in contrastive deep supervision are supervised with contrastive learning and thus they can learn the invariance to different data augmentation. As shown in previous research, these data augmentation invariance is beneficial to various downstream tasks~\cite{contrastlearningsurvey}.
In this paper, we further propose to improve knowledge distillation with contrastive deep supervision by transferring the data augmentation invariance learned by the teachers to the students. 
Denote the student model and the teacher model in knowledge distillation as  $f^\mathcal{S}$ and $f^\mathcal{T}$ respectively, the na\"ive feature-based knowledge distillation directly minimizes the distance between the backbone features of the student and the teacher, which can be formulated as 
\begin{equation}
    \label{feat_kd} \mathop{\sum}_{i=1}^{K} \| f_i^\mathcal{T}(\mathcal{X}) - f_i^\mathcal{\mathcal{S}}(\mathcal{X})\|_2.
\end{equation}

In contrast, knowledge distillation with contrastive deep supervision minimizes the distance between the embedding vectors (the output of the projection heads) of the student and the teacher, which can be formulated as
\begin{equation}
\label{ours_kd}
\begin{aligned}
     \mathcal{L_{\text{CDS for KD}}}=\mathop{\sum}_{i=1}^{K-1} \|c^{\mathcal{T}}_i(\mathcal{X})-c^{\mathcal{S}}_i(\mathcal{X})\|_2.
\end{aligned}
\end{equation} 

Now we can formulate the overall training loss of the student as 
\begin{equation}
\label{ours_kd}
\begin{aligned}
    \mathcal{L}_{\text{DCDS}} &= \mathcal{L}_{\text{CDS}} + \lambda_2 \cdot \mathcal{L_{\text{CDS for KD}}} + \lambda_3 \cdot \mathcal{L}_{\text{KL}}\left(c^{\mathcal{T}}_K\left(\mathcal{X}\right), c^{\mathcal{S}}_K\left(\mathcal{X}\right)\right),
\end{aligned}
\end{equation}
where $\lambda_2$ and $\lambda_3$ are the hyper-parameters to balance different loss items. Following previous works in deep supervision, we do not set an individual hyper-parameter for each projection head for convenience in hyper-parameter tuning. 


\subsection{Other Details and Tricks}

\noindent{\textbf{Design of Projection Heads}}
In contrastive deep supervision, several projection heads are added to the intermediate layers of neural networks during the training period. These projection heads map the backbone features into a normalized embedding space, where the contrastive learning loss is applied. As discussed in related works, the architecture of the projection head is crucial to model performance~\cite{simclr2}. 
Usually, the projection head is a non-linear projection stacked by two fully connected layers and a ReLU function. However, in contrastive deep supervision, the input feature comes from the intermediate layers instead of the final layer, and thus it is more challenging to project them properly~\cite{simclr2}.
Hence, we increase the complexity of these projection heads by adding convolutional layers before the non-linear projection.




\begin{table*}[t!]
    \caption{\label{table:cifar100_deepsupervision} Comparison experiments (top-1 accuracy / \%) with the other  deep supervision methods on CIFAR100.
    }
            \setlength{\tabcolsep}{0.2mm}{\begin{tabular}{llllllllllll}
    \toprule
    Method  &RNT18   &RNT50& RNT101   &RXT50  &RXT101&    WRN50&  WRN101&SET18&SET50&PAT18 \\
    \midrule
    Base&77.45&77.81&78.65&79.85&80.67&79.46&79.98&77.46&78.02&76.84 \\
    DSN&78.30&78.96&79.37&81.02&81.70&80.98&81.30&78.28&79.46&77.40 \\
    DKS&78.96&80.95&81.39&82.27&82.98&81.95&82.58&79.32&80.76&78.96 \\
    DHM&78.82&81.12&81.27&82.14&83.27&81.76&82.76&79.14&80.72&78.32\\
    \cellcolor{mygray}
    \textbf{Ours}&\cellcolor{mygray}\textbf{80.84}&\cellcolor{mygray}\textbf{81.31}&\cellcolor{mygray}\textbf{83.12}&\cellcolor{mygray}\textbf{82.81}&\cellcolor{mygray}\textbf{83.87}&\cellcolor{mygray}\textbf{82.28}&\cellcolor{mygray}\textbf{83.93}&\cellcolor{mygray}\textbf{80.13}&\cellcolor{mygray}\textbf{81.51}&\cellcolor{mygray}\textbf{80.76}\\
    \bottomrule
    \end{tabular}}
    \begin{center}
    \end{center}
\end{table*}

\begin{table*}[t!]
    \caption{\label{table:cifar10_deepsupervision} Comparison experiments (top-1 accuracy  / \%) with the other deep supervision methods on CIFAR10. 
    }
    
            \setlength{\tabcolsep}{0.2mm}{\begin{tabular}{llllllllllll}
    \toprule
    Method  &RNT18   &RNT50& RNT101   &RXT50  &RXT101&    WRN50&  WRN101&SET18&SET50&PAT18 \\
    \midrule
    Base&94.96&95.07&95.13&95.09&95.34&95.01&95.27&94.86&95.11&94.78\\
    DSN&95.31&95.41&95.63&95.39&95.70&95.27&95.78&95.21&95.41&95.13\\
    DKS&95.72&95.90&96.21&95.98&96.10&95.50&96.12&95.74&95.72&95.47\\
    DHM&95.61&95.87&96.04&96.10&96.27&95.62&96.31&95.59&95.77&95.38\\
    \cellcolor{mygray}\textbf{Ours}&\cellcolor{mygray}\textbf{96.49}&\cellcolor{mygray}\textbf{96.78}&\cellcolor{mygray}\textbf{97.02}&\cellcolor{mygray}\textbf{96.76}&\cellcolor{mygray}\textbf{97.05}&\cellcolor{mygray}\textbf{96.88}&\cellcolor{mygray}\textbf{97.01}&\cellcolor{mygray}\textbf{96.50}&\cellcolor{mygray}\textbf{96.73}&\cellcolor{mygray}\textbf{96.37}\\
    \bottomrule
    \end{tabular}}
    \begin{center}
    \end{center}
\end{table*}

\noindent{\textbf{Contrastive Learning}}
The proposed contrastive deep supervision is a general training framework and does not depend on a specific contrastive learning method. 
In this paper, we adopt  SimCLR~\cite{simclr} and SupCon~\cite{sup} as the contrastive learning method in most experiments. 
We argue that the performance of our method can be further improved by using better contrastive learning method.

\begin{table*}[t!]
	\caption{\label{table:imagenet_deepsupervision} Comparison with the other deep supervision methods on ImageNet.}
    \begin{center}
        \setlength{\tabcolsep}{4.0mm}{
    \begin{tabular}{lccccccccccccccccc}
      \toprule
     Metric&Model &Baseline & DSN& DKS& DHM&\textbf{Ours}\cellcolor{mygray}\\ \midrule
     \multirow{3}{*}{top-1}&RNT18&69.21&69.54&71.32&71.29&\textbf{72.85}\cellcolor{mygray}\\
     &RNT34&73.17&73.29&74.01&73.89&\textbf{76.19}\cellcolor{mygray}\\
     &RNT50&75.30&75.37&76.47&76.57&\textbf{78.25}\cellcolor{mygray}\\
     \midrule
     \multirow{3}{*}{top-5}&RNT18&89.01&88.87&89.20&90.06&\textbf{91.30}\cellcolor{mygray}\\ 
     
     &RNT34&91.24&91.30&91.87&91.66&\textbf{93.08}\cellcolor{mygray}\\ 
     &RNT50&92.20&92.49&93.60&93.24&\textbf{93.99}\cellcolor{mygray}\\
     \bottomrule
\end{tabular}}
\end{center}
\end{table*}

\noindent{\textbf{Negative Samples}} 
Previous studies show that the number of negative samples has a vital influence on the performance of contrastive learning. Accordingly, a large batch size, a momentum encoder or a memory bank is usually required~\cite{simclr,moco,byol}. In contrastive deep supervision, we do not use any of these solutions because the supervised loss ($\mathcal{L}_{\text{CE}}$ in Equation~\ref{loss:ours}) is enough to prevent contrastive learning from converging to the collapsing solutions.

\begin{table}[t!]
        \begin{center}
        \caption{\label{tab:object} Experiments on different object detection models on COCO2017. ResNet50 models are pre-trained on ImageNet with different deep supervision methods and then utilized as the backbones of these detectors.}
        \setlength{\tabcolsep}{3.0mm}{\begin{tabular}{llllll}
          \toprule
          Model   & Method&\textbf{AP}&\textbf{AP$_{S}$}&\textbf{AP$_{M}$}&\textbf{AP$_{L}$} \\ 
          
         \midrule
         \multirow{5}{*}{Faster RCNN}&  Baseline&37.4 &21.2 &41.0&48.1\\ 
          &DSN &37.3$_{-0.1}$&  21.0$_{-0.2}$&40.8$_{-0.2}$&48.3$_{-0.2}$\\ 
          &DKS &37.5$_{+0.1}$&  21.2$_{+0.0}$&41.5$_{+0.5}$&47.6$_{-0.5}$\\ 
          &DHM &37.6$_{+0.2}$&  21.3$_{+0.1}$&41.3$_{+0.3}$&48.2$_{+0.1}$ \\ 
          &\textbf{Ours} \cellcolor{mygray}&\cellcolor{mygray}\textbf{38.3}$_{\mathbf{+0.9}}$&\textbf{21.6$_{\mathbf{+0.4}}$} \cellcolor{mygray}&\textbf{42.0$_{\mathbf{+1.0}}$} \cellcolor{mygray}&\textbf{50.1}$_{\mathbf{+2.0}}$ \cellcolor{mygray}\\ 
         \midrule
        \multirow{5}{*}{RetinaNet} & Baseline&36.5 &20.4 &40.3&48.1\\ 
          &DSN &36.3$_{-0.2}$&20.1$_{-0.3}$&40.0$_{-0.3}$&48.1$_{0.0}$  \\ 
          &DKS &36.7$_{+0.2}$& 20.1$_{-0.3}$&40.9$_{+0.6}$&48.2$_{+0.1}$ \\ 
          &DHM &36.7$_{+0.2}$&20.0$_{-0.4}$&40.7$_{+0.4}$&48.5$_{+0.4}$  \\ 
          &\textbf{Ours} \cellcolor{mygray}&\textbf{37.3$_{\mathbf{+0.8}}$}\cellcolor{mygray}&\textbf{21.2$_{\mathbf{+0.8}}$} \cellcolor{mygray}&\textbf{41.0$_{\mathbf{+0.7}}$} \cellcolor{mygray} &\textbf{47.9$_{\mathbf{-0.2}}$} \cellcolor{mygray}\\ 
         \bottomrule
        \end{tabular}}
        \end{center}
    \end{table}

        \begin{table*}[t!]
            \caption{\label{tab:fine-grained} Comparison (top-1 acc. / \%) with deep supervision methods with ResNet50 for fine-grained classification. Models are trained from scratch. }
            \centering
            \setlength{\tabcolsep}{2.5mm}{
            \begin{tabular}{l l l llll}
            \toprule 
            Method & CUB & Cars &Flowers&  Dogs& Aircrafts\\
            \midrule
            Baseline&60.65&79.86&87.52&64.00&74.07\\
            DSN &62.37$_{+1.72}$&81.04$_{+1.18}$&88.54$_{+1.02}$&66.32$_{+2.32}$&74.49$_{+0.42}$\\
            DKS&63.59$_{+2.94}$&81.52$_{+1.66}$&88.94$_
            {+0.40}$&68.31$_{+4.31}$&75.07$_{+1.00}$\\
            DHM&64.01$_{+3.36}$&81.49$_{+1.63}$&89.03$_{+1.51}$&68.38$_{+4.38}$&75.00$_{+0.93}$\\
            \cellcolor{mygray}\textbf{Ours}&\cellcolor{mygray}\textbf{64.65}$_{\mathbf{+4.00}}$&\cellcolor{mygray}\textbf{82.07}$_{\mathbf{+2.21}}$&\cellcolor{mygray}\textbf{89.26}$_{\mathbf{+1.74}}$&\cellcolor{mygray}\textbf{69.02}$_{\mathbf{+5.02}}$&\cellcolor{mygray}\textbf{75.43}$_{\mathbf{+1.36}}$\\
            \bottomrule
            \end{tabular}}
            \end{table*}

            \begin{table*}[t!]
                    \caption{\label{tab:fine-grained} Comparison (top-1 acc. \%) with deep supervision methods with ResNet50 for fine-grained classification. Models are finetuned from ImageNet pre-trained weights.}
                    \centering
                    \setlength{\tabcolsep}{2.5mm}{
                    \begin{tabular}{l l l llll}
                    \toprule 
                    Method & CUB&Cars &Flowers&  Dogs&  Aircrafts\\
                    \midrule
                    Baseline&78.50&90.25&97.68&76.47&87.43\\
                    DSN&80.14$_{+1.64}$&91.32$_{+1.07}$&98.64$_{+0.96}$&77.21$_{+0.74}$&89.31$_{+1.88}$\\
                    DKS&81.34$_{+2.84}$&92.54$_{+2.29}$&99.01$_{+1.33}$&78.32$_{+1.85}$&89.20$_{+1.77}$\\
                    DHM&81.27$_{+2.77}$&92.31$_{+2.06}$&98.84$_{+1.16}$&78.20$_{+1.73}$&89.57$_{+2.14}$\\
                    \cellcolor{mygray}\textbf{Ours}&\cellcolor{mygray}\textbf{82.10}$_{\mathbf{+3.60}}$&\cellcolor{mygray}\textbf{92.90}$_{\mathbf{+2.65}}$&\cellcolor{mygray}\textbf{99.39}$_{\mathbf{+1.71}}$&\cellcolor{mygray}\textbf{80.99}$_{\mathbf{+4.52}}$&\cellcolor{mygray}\textbf{90.52}$_{\mathbf{+3.09}}$\\
                    \bottomrule
                    \end{tabular}}
                    \end{table*}

\section{Experiment}
\subsection{Experiment Setting}
\noindent{\textbf{Common Image Classification}}
For common image classification, our method has been evaluated on three datasets, including CIFAR10, CIFAR100 and ImageNet~\cite{cifar,imagenet} with  kinds of neural networks including ResNet (RNT), ResNeXt (RXT), Wide ResNet (WRN), SENet (SET), PreAct ResNet (PAT), MobileNetv1, MobileNetv2, ShuffleNetv1 and ShuffleNetv2~\cite{resnet,ResNeXt,wideresnet,senet,preactresnet,mobilenets,mobilenetv2,shufflenetv1}. 


\noindent{\textbf{Fine-grained Image Classification} }
For fine-grained image classification, our method has been evaluated on five popular datasets, including CUB200-2011~\cite{fgvc_cub}, Stanford Cars~\cite{fgvc_car}, Oxford Flowers~\cite{fgvc_flower}, Stanford Dogs~\cite{fgvc_dog} and FGVC Aircraft~\cite{fgvc_aircraft}.
ResNet50 is utilized as the classifier for all the experiments.

\noindent{\textbf{Object Detection}} For object detection, our method has been evaluated on MS COCO2017~\cite{mscoco} with Faster RCNN and RetinaNet by MMdetection~\cite{mmdetection}.

\noindent{\textbf{Semi-supervised Learning}} Semi-supervised learning experiments have been conducted on CIFAR100, CIFAR10 with ResNet18. For each dataset, we have evaluated our method with 10\%, 20\%, 30\% and 40\% labels.

\begin{table*}[t!]
    \caption{\label{table:imagenet_kd} Comparison experiments (top-1 and top-5 accuracy  / \%) with the other eight knowledge distillation methods on ImageNet with ResNet. Numbers in bold indicate the highest.
     Results marked with $^\dag$ come from the paper of SSKD~\cite{kd_sskd}. 
    }


    \begin{center}
        \setlength{\tabcolsep}{0.9mm}{
    \begin{tabular}{lccccccccccccccccc}
      \toprule
     Metric&Model &Base & KD& AT& RKD& SP  &CRD&CC$^\dag$&OKD$^\dag$&SSKD$^\dag$&\textbf{Ours}\cellcolor{mygray}\\ \midrule
     \multirow{3}{*}{top-1}&RNT18&69.21&70.52&70.74&70.63&70.61&71.07&69.96&70.55&71.62&\textbf{73.23}\cellcolor{mygray}\\\
     &RNT34 &73.17&74.44&74.69&74.61&74.60&74.99&--&--&--&\cellcolor{mygray}\textbf{76.65}\\
     &RNT50&75.30&76.62&76.79&76.92&76.88&77.21&--&--&--&\cellcolor{mygray}\textbf{78.68}\\
     \midrule
     \multirow{3}{*}{top-5}&RNT18&89.01&89.88&90.00&89.71&89.80&91.06&89.17&89.59&90.67&\textbf{91.56}\cellcolor{mygray}\\ 
     
     &RNT34&91.24&92.07&92.18&92.14&92.10&92.58&--&--&--&\textbf{93.38}\cellcolor{mygray}\\ 
     &RNT50 &92.20&93.36&93.51&93.60&93.58&93.88&--&--&--&\textbf{94.42}\cellcolor{mygray}\\
     \bottomrule
\end{tabular}}
\end{center}
\end{table*}

\begin{figure}[htbp]
        \begin{minipage}[t]{0.62\textwidth}
        \centering
    \includegraphics[height=3.0cm]{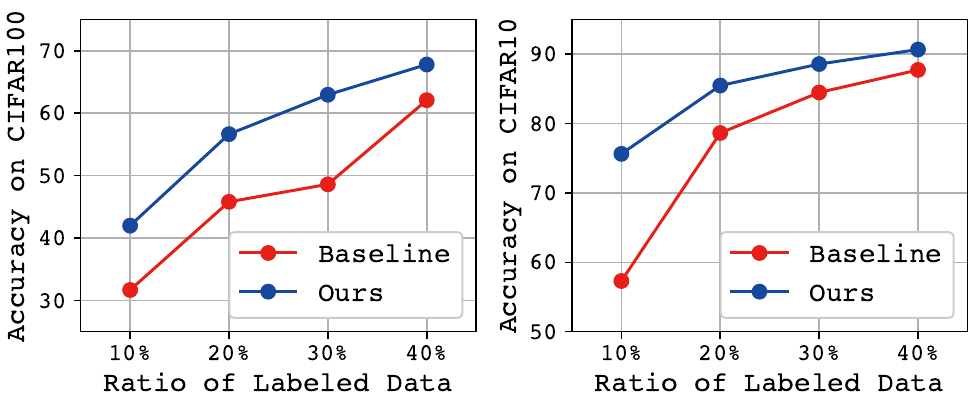}
    \caption{Experimental results of semi-supervised training on CIFAR100 and CIFAR10 with ResNet18.}
    \label{fig:semi}
        \end{minipage}
        \begin{minipage}[t]{0.02\textwidth}
        ~~
        \end{minipage}
        \begin{minipage}[t]{0.32\textwidth}
        \centering
    \includegraphics[height=3.2cm]{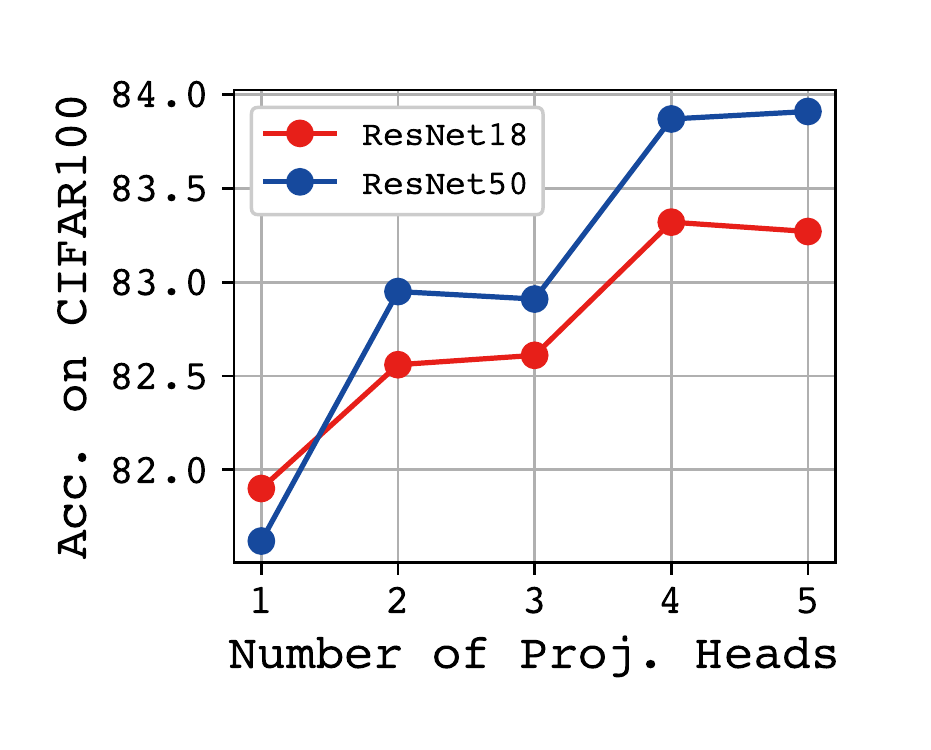}
    \caption{Influence from the number of projection heads.}
    \label{fig:num}
    
        \end{minipage}
\end{figure}
\noindent{\textbf{Comparison Methods}}
Three previous deep supervision methods are utilized for comparison, including DSN~\cite{deeplysupervisednet}, DKS~\cite{dsk} and DHM~\cite{li2020dynamic}. 
In knowledge distillation experiments, we have evaluated our method with nine knowledge distillation methods, including KD~\cite{distill_hinton}, FitNet~\cite{fitnets}, AT~\cite{attentiondistillation}, RKD~\cite{relational_kd}, SP~\cite{relational_kd2} and CRD~\cite{kd_crd}. Besides, we also cite results on ImageNet of CC~\cite{kd_cc}, OKD~\cite{kd_online}, and SSKD~\cite{kd_sskd} from the paper of SSKD.


\begin{table*}[h!]
    \caption{\label{table:cifar100} Comparison with the other knowledge distillation methods on CIFAR. 
    }
    
    \begin{center}
    \setlength{\tabcolsep}{1.75mm}{
    \begin{tabular}{llllllllllllllllllllll}
     \multicolumn{9}{c}{\texttt{\textbf{CIFAR100}}} \\ 
     \toprule
     Model&Base & KD & FitNet & AT & RKD& SP&CRD&\textbf{Ours}\textbf{} \\ \cmidrule{1-9} 
     ResNet18&77.45&78.68&78.15&78.09&78.21&78.19   &81.41&\textbf{83.31}   \\
     ResNet50&77.81&79.19&78.42&78.34&78.94&78.81   &82.45&\textbf{83.53}  \\
     
     ResNet101&78.65&80.40&80.78&80.97&81.24&80.94  &82.57&\textbf{84.80}  \\
     
     ResNeXt50&79.85&81.41&82.67&82.59&83.71&82.67  &83.41&\textbf{84.41}  \\
     ResNeXt101&80.67&82.03&82.51&82.43&83.01&82.64 &84.50&\textbf{85.37}  \\

     WRNet50&79.46&81.02&81.29&81.16&82.06&82.07    &82.94&\textbf{84.27}  \\
     WRNet101&79.98&81.82&82.07&82.16&82.54&82.49   &83.07&\textbf{85.04}    \\

     SENet18 &77.46&78.92&79.09&79.15&79.41&79.31   &81.22&\textbf{82.68}   \\
     SENet50 &78.02&79.78&80.13&80.45&80.69&80.71   &81.79&\textbf{83.36}   \\
     SENet101 &78.92&80.31&80.54&80.53&80.74&80.52  &82.75&\textbf{84.15}   \\
    
     MobileNetV1 &68.32&70.04&70.25&70.17&70.89&70.19   &72.68&\textbf{73.79}   \\
     MobileNetV2 &69.34&70.58&70.64&70.51&70.83&70.68   &71.82&\textbf{72.61}    \\
     
     ShuffleNetV1 &72.46&74.08&74.19&74.11&74.56&74.68  &75.11&\textbf{75.77}   \\
     ShuffleNetV2 &72.81&74.39&74.47&74.51&74.82&74.67  &75.62&\textbf{76.11}   \\
     PreActNet18&76.84&78.25&78.34&78.67&79.01&79.12    &81.62&\textbf{82.83}  \\ 
     PreActNet50&77.31&79.04&79.27&79.54&79.82&79.76    &81.27&\textbf{83.42}   \\ 
     \bottomrule
\end{tabular}
\\
\begin{tabular}{llllllllllllllllllllll}
   \multicolumn{9}{c}{\texttt{\textbf{CIFAR10}}} \\ 
   \toprule
   Model&Base & KD & FitNet & AT & RKD& SP&CRD&\textbf{Ours}\textbf{}\\ \midrule
   ResNet18   &94.96&95.24&95.31&95.26&95.31&95.27   &95.81&\textbf{96.84} \\
   ResNet50   &95.07&95.31&95.45&95.47&95.33&95.29   &96.21&\textbf{97.08}\\
   
   ResNet101&95.13&95.39&95.71&95.49&95.43&95.18  &96.37&\textbf{97.40}\\
   
   ResNeXt50&95.09&95.27&95.36&95.68&95.59&95.37  &96.49&\textbf{97.15}\\
   ResNeXt101&95.34&95.68&95.92&95.78&95.81&95.38 &96.51& \textbf{97.40}\\

   WRNet50&95.01&95.34&95.38&95.34&95.61&95.73    &96.17&\textbf{97.37}\\
   WRNet101&95.27&95.51&95.48&95.71&95.99&95.82   &96.34&\textbf{97.39} \\

   SENet18 &94.86&95.21&95.30&95.47&95.34&95.41   &96.00&\textbf{96.96}\\
   SENet50 &95.11&95.39&95.44&95.64&95.57&95.47   &96.21&\textbf{97.19}\\
   SENet101 &95.30&95.64&95.81&95.78&95.81&95.77  &96.19&\textbf{97.36}\\
  
   MobileNetV1 &90.24&91.27&92.59&92.87&93.01&92.90   &93.27&\textbf{93.94} \\
   MobileNetV2 &90.76&91.09&91.57&91.75&91.82&91.83   &92.17&\textbf{92.87} \\
   
   ShuffleNetV1 &91.57&91.99&92.30&92.19&92.47&92.38  &93.08&\textbf{94.04}\\
   ShuffleNetV2&91.19&91.87&92.23&92.41&92.30&92.54  &92.90&\textbf{93.16}\\
   
   PreActNet18&94.78&95.08&95.28&95.39&95.51&95.69    &96.07&\textbf{96.70} \\ 
   PreActNet50&94.89&95.21&95.57&95.49&95.37&95.48    &96.11&\textbf{96.93}\\ 

   \bottomrule
\end{tabular}}
\end{center}

\end{table*}

\subsection{Experimental Results}
\noindent{\textbf{Image Classification}}
Experimental results on CIFAR100, CIFAR10 and ImageNet are shown in Table~\ref{table:cifar100_deepsupervision}, Table~\ref{table:cifar10_deepsupervision} and Table~\ref{table:imagenet_deepsupervision}, respectively.
It is observed that: \textbf{(a)} Our method achieves 3.44\% and 1.70\% top-1 accuracy improvements on CIFAR100 and CIFAR10 on average, respectively. 
It consistently outperforms the second-best deep supervision method by 1.05\% and 0.90\% on the two datasets, respectively.
\textbf{(b)} On ImageNet, contrastive deep supervision leads to 3.64\%, 3.02\% and 2.95\% top-1 accuracy improvements on ResNet18, ResNet34 and ResNet50, respectively. On average, it outperforms the baseline and the second-best method by 3.20\% and 1.83\% top-1 accuracy, respectively.

\noindent{\textbf{Object Detection}} 
Table~\ref{tab:object} shows the the performance of our method on  object detection. In these experiments,
We firstly pretrain the ResNets on ImageNet with standard training (Baseline), three deep supervision methods, and our method, and then finetuning them as the backbone for object detection models, including RetinaNet and Faster RCNN on COCO2017 datasets. It is observed that with backbones pre-trained with our method, there are 0.9 and 0.8 AP improvements on Faster RCNN and RetinaNet respectively, which outperforms the second-best method by 0.6 AP, indicating that the representation learned with our method are more beneficial to downstream tasks.

\noindent{\textbf{Fine-grained Image Classification}}
Experiments on fine-grained image classification are shown in Table~
\ref{tab:fine-grained}. It is observed that: \textbf{(a)} Contrastive deep supervision leads to consistent and significant accuracy improvements on the five datasets. On average, it leads to 3.80\%, 2.43\%, 1.73\%, 4.77\% and 2.25\% accuracy improvements on the five datasets, respectively. \textbf{(b)} Besides, the benefits of our method in \emph{``finetuning from ImageNet''} and \emph{``training from scratch''} are very similar (except on Aircraft), which indicates that the effectiveness of our method is consistent in different training settings.

\noindent{\textbf{Semi-supervised Learning}}
Experiments on semi-supervised learning with ResNet18 on CIFAR10 and CIFAR100 are shown in Figure~\ref{fig:semi}. It is observed that: \textbf{(a)} Our method leads to consistent accuracy improvements at all the ratios of labeled data. \textbf{(b)} The benefits of our method become larger when there is less labeled data, which indicates that our method is effective in using the unlabeled data to optimize the intermediate layers.

\noindent{\textbf{Knowledge Distillation}}
Knowledge distillation experiments on ImageNet and CIFAR are shown in Table~\ref{table:imagenet_kd} and Table~\ref{table:cifar100}, respectively.
It is observed that: \textbf{(a)} Our method achieves 5.07\% and 2.20\% top-1 accuracy improvements on CIFAR100 and CIFAR10 on average, outperforming the second-best KD method by 1.40\% and 0.87\% on the two datasets, respectively.
\textbf{(b)} The similar results can also be observed in ImageNet experiments. Our method leads to 4.02\%/2.55\%, 3.48\%/2.14\% and 3.38\%/2.22\% top-1/top-5 accuracy improvements on ResNet18, ResNet34 and ResNet50, respectively. On average, it outperforms the baseline and the second-best method by 3.62\% and 1.76\%  top-1 accuracy, respectively.  

\section{Discussion}
\subsection{Contrastive Deep Supervision as a Regularizer} 
\begin{figure}
    \begin{minipage}[t]{0.45\textwidth}

        \centering
    \includegraphics[width=5.7cm]{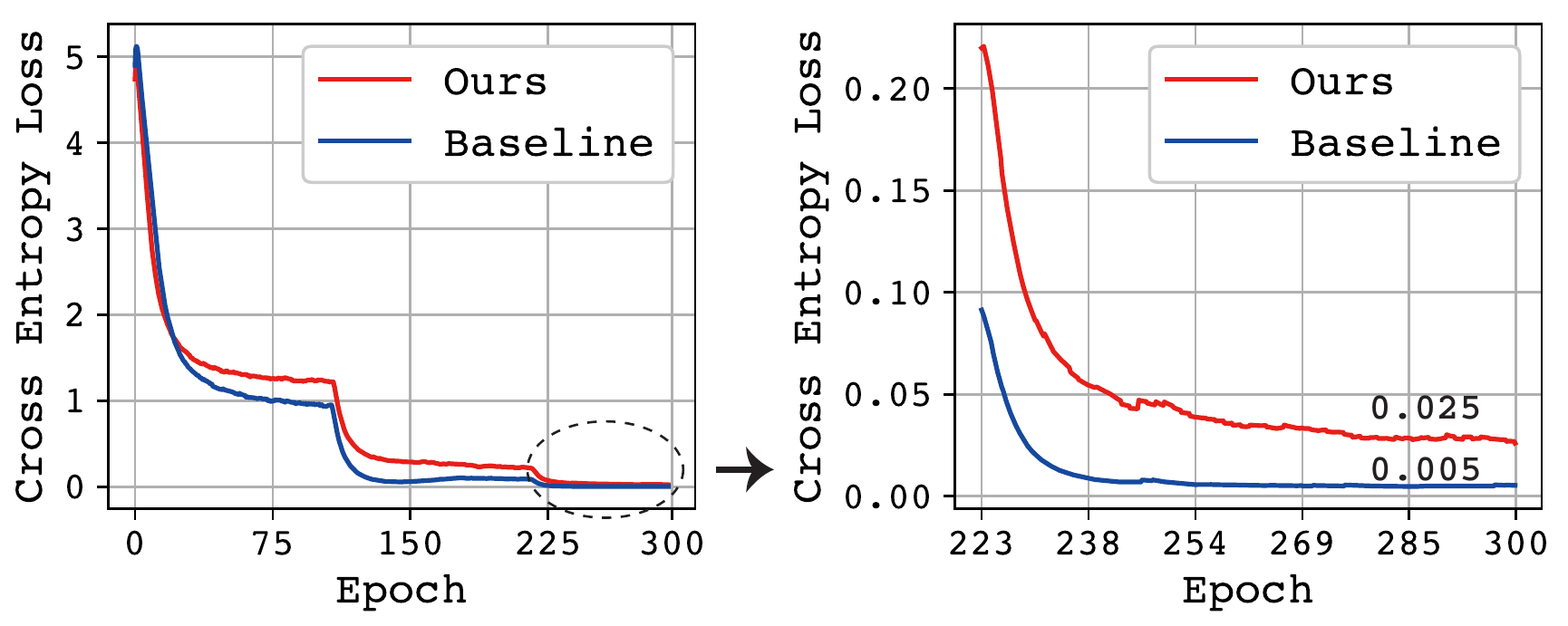}
    
    \caption{Comparison on the cross entropy loss between predicted results and labels during the training period. Note that our method also leads to better accuracy (80.84\% vs 77.45\%)}
    \label{fig:loss}
    \end{minipage}
    \begin{minipage}[t]{0.02\textwidth}
        ~~
    \end{minipage}
    \begin{minipage}[t]{0.47\textwidth}

        \centering
        \includegraphics[width=5.7cm]{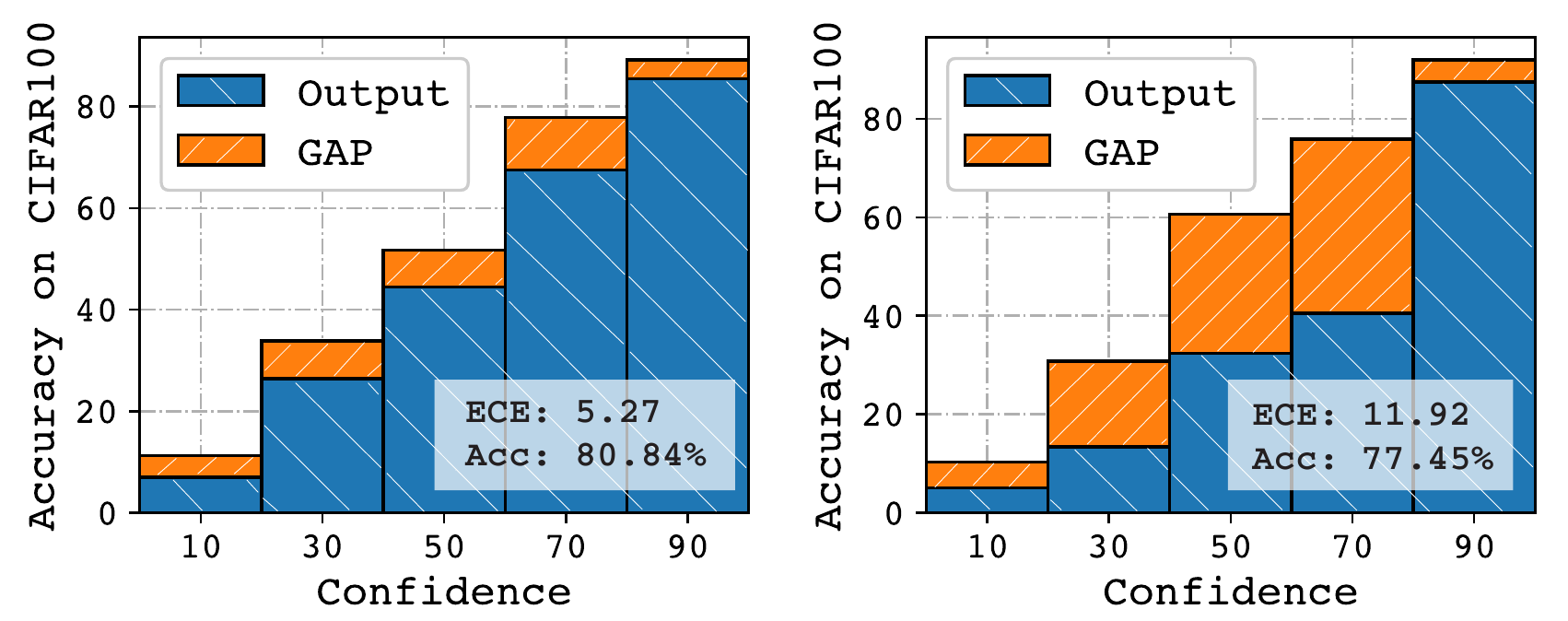}
        
        \caption{Comparison on  reliability diagrams. ``GAP'' indicates the difference between confidence and accuracy. ``Output'' indicates accuracy.  ECE: Expected Calibrated Error (lower is better).}
        \label{fig:ece}
    \end{minipage}
\end{figure}

\noindent{\textbf{Loss Curves}} Regularization methods in deep learning are usually utilized to avoid model overfitting
by introducing additional penalties or loss. In this subsection, we show that the contrastive learning loss introduced by our method in the intermediate layers works as a regularizer. Figure~\ref{fig:loss}  shows the cross entropy loss between predicted results and labels during the training period from two ResNet18 models trained by the standard method and our method, respectively. It is observed that at most of epochs, the baseline model has lower cross entropy loss than our model. When both models are converged (epoch 280-300), the baseline model has only 0.005 loss while our model still has 0.025 loss. These observations indicate that there is serve overfitting in the baseline model while deep contrastive supervision can alleviate overfitting and thus improve the accuracy.


\noindent{\textbf{Uncertainty Estimation}} Besides, the comparison on expected calibrated error (ECE) of models trained with the standard method and our method has been shown in Figure~\ref{fig:ece}. A lower ECE indicates that the predicted probability of a neural network estimates representative of the
true correctness likelihood better~\cite{guo2017calibration}.
It is observed that compared with the baseline model, our method leads to a lower ECE, indicating better uncertainty estimation and interpretability.

\begin{table}[t!]
    \caption{\label{contra}
    Comparison between our method and contrastive learning methods with ResNet50 on ImageNet.
    Baseline$^1-2$: Two baselines trained with and without AutoAugmentation~\cite{AutoAugmentLA}. 
    SupCon$^{1-3}$: Three models trained by supervised contrastive learning with different hyper-parameters. 
    BYOL: ResNet50 unsupervisedly pre-trained by 1000 epochs and then supervisedly finetuned. BYOL+DSN: ResNet50 pretrained with BYOL and then finetuned with deep supervision. 
    Ours$^{1,3}$: ResNet50 trained with contrastive deep supervision in different settings. Ours$^2$: ResNet50 trained with contrastive deep supervision+ knowledge distillation. 
    }
    \begin{center}
    \setlength{\tabcolsep}{4.5mm}{\begin{tabular}{l l l c c}
    \toprule
         Method& Batchsize & Epoch & AutoAug & top-1 acc. (\%)  \\
         \midrule
         Baseline$^1$& 256& 90 & $\times$ & 75.3\\
         Baseline$^2$& 4096 & 270 & $\checkmark$ & 77.6\\
         SupCon$^1$&6144 & 350 & $\checkmark$ & 78.7\\
         SupCon$^2$&512 & 350 & $\checkmark$ & 74.5\\
         SupCon$^3$&6144 & 100 & $\checkmark$ & 77.0\\
         BYOL&1024 & 1080 & $\times$ & 77.7\\
         BYOL+DSN&1024 & 1080 & $\times$ & 78.2\\
         Ours$^1$& 256&90&$\times$&78.3\\
         Ours$^2$& 256&90&$\times$&78.7\\
         Ours$^3$\cellcolor{mygray}& 256\cellcolor{mygray}&350\cellcolor{mygray}&$\checkmark$\cellcolor{mygray}&79.8\cellcolor{mygray}\\
        \bottomrule
            
    \end{tabular}}
    \end{center}

    \end{table}

\subsection{Comparison with Contrastive Learning}
Comparison between our method and two ``pretrain \& finetune''  contrastive learning methods is shown in Table~\ref{contra}. 
It is observed that without a large batch size and the advanced data augmentation policy (AutoAugment), contrastive deep supervision (Ours$^1$) with only 25\% training time achieves 0.4\% lower accuracy than SupCon$^3$. Besides, contrastive deep supervision with the same training time and data augmentation (Ours$^3$) achieves 1.1\% and 1.6\% higher accuracy than SupCon$^3$ and BYOL+DSN, respectively, which demonstrates the advantage of our method over the traditional contrastive learning methods.

\subsection{Ablation Study on Knowledge Distillation}
The main difference between the na\"ive feature distillation and feature distillation with our contrastive deep supervision is \emph{``what to distill''}. Na\"ive feature distillation distills the backbone features while our method distills the embedding learned by contrastive deep supervision. To further demonstrate its effectiveness, we have trained a ResNet50 model on CIFAR100 with both contrastive deep supervision and distillation on backbone features. Experimental results show this model achieves 82.26\% accuracy, which is 1.27\% lower than distilling the embedding. These results demonstrate that distilling the  embedding learned by contrastive deep supervision is more beneficial. 

\subsection{Sensitivity Study}

\noindent{\textbf{Where to Apply Projection Heads}}
We study the influence from the position of projection heads with the following four schemes: (1)\emph{uniform scheme} - applying projection heads into different depths uniformly; (2) \emph{downsampling scheme} - applying projection heds into the layers before downsampling; (3) \emph{shallow scheme} - applying projection heads into only the shallower layers; (4) \emph{deep scheme} - applying projection heads to only the deeper layers; Experimental results on CIFAR100 with ResNet50 show that the four schemes achieves 81.23\%, 81.31\%, 81.07\% and 80.99\% accuracy, respectively. It is observed that both \emph{uniform} and \emph{downsampling} schemes leads to excellent performance, indicating our method is not sensitive to where to apply projection heads.

\noindent{\textbf{The Number of Projection Heads}}
We have studied the influence from the number of projection heads in Figure~\ref{fig:num}.
It is observed that when there are less than five projection heads, more projection heads tend to achieve better performance. The fifth projection head does not leads to more accuracy improvements.

\section{Conclusion}
This paper proposes \emph{contrastive deep supervision}, a novel training methodology that directly optimizes the intermediate layers of deep neural networks with contrastive learning. It enables the neural network to learn better visual representation without additional computation and storage in inference.
Experiments on nine datasets with eleven neural networks have demonstrated its effectiveness in general image classification, fine-grained image classification and object detection for traditional supervised learning, semi-supervised learning and knowledge distillation. It outperforms the previous deep supervision methods, knowledge distillation methods, and contrastive learning methods by a clear margin.
Besides, we also show that contrastive deep supervision works as a regularizer to prevent models from overfitting, and thus leads to better uncertainty estimation. 
\clearpage
%
%
\bibliographystyle{splncs04}
\bibliography{egbib}
\end{document}